\documentclass[preprint,12pt]{elsarticle}

\usepackage{amssymb}
\usepackage{graphicx} 
\usepackage{comment}
\usepackage{enumerate}
\usepackage[shortlabels]{enumitem}
\usepackage{subcaption}
\usepackage{amsmath,nccmath}
\usepackage{acronym}
\usepackage{booktabs}
\usepackage{xspace}
\usepackage{color}
\usepackage{hyperref}

\newcommand{\RQ}[2]{
    \begin{description}[topsep=0pt,nosep=0pt, leftmargin=0.75cm, noitemsep,nolistsep]
    \phantomsection\label{section:setup:rq#1}
    \item[RQ#1] #2
    \end{description}
}
\newcommand{\RQRef}[1]{\textbf{\hyperref[section:setup:rq#1]{RQ#1}}}

\acrodef{CRS}[CRS]{Conversational Recommender System}

\newcommand{\jie}[1]{\textcolor{black}{#1}}

\journal{Elsevier}

\begin{document}

\begin{frontmatter}



\title{Multi-Type Context-Aware Conversational Recommender Systems via Mixture-of-Experts}

\author[inst1]{Jie Zou}
\author[inst1]{Cheng Lin}

\affiliation[inst1]{organization={ School of Computer Science and Engineering},
            addressline={University of Electronic Science and Technology of China}, 
            city={Chengdu},
            postcode={611731},
            country={China}}
\author[inst3]{Weikang Guo}
\affiliation[inst3]{organization={School of Management Science and Engineering},
            addressline={Southwestern University of Finance and Economics}, 
            city={Chengdu},
            postcode={611130},
            country={China}}
            
\author[inst2]{Zheng Wang}
\author[inst1]{Jiwei Wei}
\author[inst1]{Yang Yang}
\author[inst1,inst2]{Heng Tao Shen}

\affiliation[inst2]{organization={School of Computer Science and Technology}, 
            addressline={Tongji University}, 
            city={Shanghai},
            postcode={200092}, 
            country={China}}
\begin{abstract}
Conversational recommender systems enable natural language conversations and thus lead to a more engaging and effective recommendation scenario. As the conversations for recommender systems usually contain limited contextual information, many existing conversational recommender systems incorporate external sources to enrich the contextual information. However, how to combine different types of contextual information is still a challenge. In this paper, we propose a multi-type context-aware conversational recommender system, called MCCRS, effectively fusing multi-type contextual information via mixture-of-experts to improve conversational recommender systems. MCCRS incorporates both structured information and unstructured information, including the structured knowledge graph, unstructured conversation history, and unstructured item reviews. It consists of several experts, with each expert specialized in a particular domain (i.e., one specific contextual information). Multiple experts are then coordinated by a ChairBot to generate the final results. Our proposed MCCRS model takes advantage of different contextual information and the specialization of different experts followed by a ChairBot breaks the model bottleneck on a single contextual information. Experimental results demonstrate that our proposed MCCRS method 
achieves significantly higher performance compared to existing baselines.

\end{abstract}



\end{frontmatter}

\section{Introduction}
Conversational recommender systems have emerged as a prominent research topic~\cite{gao2021advances, zou2022improving-sigir} in recent years. Unlike traditional recommender systems that mainly depend on historical data and user behavior, conversational recommender systems leverage the power of natural language conversations, so as to provide personalized and context-aware recommendations. With the interaction through conversations, users are supposed to experience a more engaging and effective recommendation. 

As conversational recommender systems are gaining significant research interest, numerous approaches have been introduced to advance the field~\citep{chen2019towards,zhou2020improving,sarkar2020suggest,CRWalker,zhou2020towards,wang2021conversation,wang2022recindial,chen2021knowledge,yang2021improving,zhang2021kecrs,lu2021revcore}. Given that the conversations for recommender systems are usually short and contain limited contextual information, incorporating external sources to enrich the contextual information is soon become common in many existing studies~\cite{zhou2022c2,zhou2020improving}. Among these external sources, the most frequently used ones are knowledge graphs~\cite{ren2022variational,chen2019towards} and item reviews~\cite{zhou2022c2}. However, how to combine different types of contextual information is still challenging. 

\jie{One of the challenges is the heterogeneity of different external data. For instance, knowledge graphs are structured data while conversations and item reviews are unstructured data. There is a semantic gap between multi-type and heterogeneous data. Furthermore, different external data often belong to different semantic spaces and thus are difficult to align. One possible solution to fuse different external data is contrastive learning \citep{zhou2022c2, won2023cross}. However, contrastive learning requires the same entries contained in all different external data to calculate the contrastive loss. This is not always fulfilled.} 

In this work, we propose a \textbf{M}ulti-type \textbf{C}ontext-aware \textbf{C}onversational \textbf{R}ecommender \textbf{S}ystem, called MCCRS, to effectively fuse multi-type contextual information via mixture-of-experts so as to improve conversational recommender systems. 
MCCRS incorporates both structured information and 
unstructured information, including the structured knowledge graph, unstructured conversation history, and unstructured item reviews. We carefully design our model as a mixture-of-experts structure, which consists of a ChairBot and several experts. Each expert specializes in a particular domain, i.e., one external source. The ChairBot coordinates multiple experts and generates the final results. \jie{Compared with previous approaches, the advantages of MCCRS are at least four-fold: First, our MCCRS model relieves the challenge of data heterogeneity and semantic fusion, as well as relaxes the limitation of contrastive learning. Second, our MCCRS model takes advantage of different contextual information and the specialization of different experts followed by a ChairBot breaks the model bottleneck on a single contextual information. Third, with each expert specialized in a particular domain, it is more easily traceable and analyzes who is responsible when the model generates a good or bad result. Fourth, our MCCRS model is more easily extendable as additional experts can be added easily to incorporate more additional external sources.}

In summary, the main contributions of this work are threefold:
\begin{itemize}
    \item We propose a new method for conversational recommender systems, named MCCRS. 
    \item We propose a new mixture-of-experts framework consisting of a ChairBot and multi-experts to effectively leverage multi-type contextual information for conversational recommender systems. 
    \item The extensive experiments show that our proposed model MCCRS significantly enhances performance compared to various state-of-the-art (SOTA) baselines. 
\end{itemize}
To the best of our knowledge, MCCRS is the first model that devises a ChairBot and multi-experts for conversational recommender systems.

\section{Related Work}
\label{sec:rel}
Conversational recommender systems utilize human-like natural language to deliver personalized and engaging recommendations through conversational interfaces like chatbots and intelligence assistants\citep{gao2021advances,jannach2021survey,shen2023towards,li2022self,zhang2024goal}. In recent years, it has attracted considerable attention, driven by the rapid evolution of dialog systems. 

In general, conversational recommender systems can be classified into two primary classes~\citep{zhang2021kecrs,zhou2020towards,he2024tut4crs,zou2024knowledge}, including \textit{anchor-based conversational recommender systems} \citep{zou2020towardsb} and \textit{dialog-based conversational recommender systems} \citep{zou2022improving-sigir,zhang2024towards}. 

\paragraph{Anchor-based Conversational Recommender Systems} This category of conversational recommender systems~\citep{zou2020towardsb} is in the form of ``system asks – user response'' mode.  In this context, the systems ask questions, prompting users to provide feedback, subsequently utilizing this user-provided information to refine their recommendations. For building effective anchor-based conversational recommender systems, recent research has introduced a variety of ways for generating anchors to characterize items, including intent slots (e.g., item aspects and facets)\citep{zhang2018towards,sun2018conversational}, entities \citep{zou2020towardsb}, topics \citep{christakopoulou2018q}, and attributes \citep{lei2020estimation,lei2020interactive,hu2022learning,zhang2022multiple}. These anchors are usually collected to construct a predefined question pool. Based on the constructed question pool, a line of work adopt multi-armed bandit \citep{christakopoulou2016towards, christakopoulou2018q, li2021seamlessly, dai2024conversational}, reinforcement learning \citep{lei2020estimation}, or greedy strategies \citep{christakopoulou2016towards, zou2022learning} to select appropriate questions to simulate multi-turn interactions to interact with users. By doing so, user feedback through conversational interactions is leveraged to optimize recommendation policies. In addition to selecting appropriate questions, deep reinforcement learning techniques are applied to train policy networks to determine whether to recommend items or ask a question in a turn \citep{lei2020interactive, hu2022learning}. This category of studies typically makes use of predefined dialog templates to construct conversations, with a primary focus on how to offer recommendations within the fewest possible conversation turns \citep{zou2020towardsb}. In other words, they do not focus on the modeling of natural language conversations \citep{chen2019towards}. 

\paragraph{Dialog-based Conversational Recommender Systems} The other category, known as dialog-based conversational recommender system~\citep{liu2020towards, liao2019deep, chen2021knowledge, liang2021learning, deng2022unified, pan2022keyword}, is based on human-generated dialogs. Unlike anchor-based conversational recommender systems simulating conversations based on extracted anchor text, dialog-based conversational recommender systems concentrate on generating human-like responses while making accurate recommendations. In the early stages, as there are no publicly available datasets for dialog-based conversational recommender systems, \citet{li2018towards} propose a conversational recommender system dataset in the movie domain, named ReDial. Subsequently, \citet{hayati2020inspired} propose the INSPIRED dataset tailored for conversational recommender systems with sociable recommendation strategies. 
Dialogues in these datasets typically encompass items and entities that can be linked to knowledge bases like DBpedia. Consequently, many existing studies incorporate external knowledge graphs to augment conversation information and accurately capture user preference~\citep{wang2021conversation, chen2019towards, zhou2020improving, zhang2021kecrs, zhang2021kers, ren2024explicit, li2023trea, qiu2024knowledge}. These studies typically employ graph neural networks to encode the knowledge graph and user preferences, alongside a dialog management module to guide the conversation. For example, \citet{chen2019towards} introduce the entity-oriented knowledge graph (DBpedia~\citep{bizer2009dbpedia}), to discern user intentions. They employ Relational Graph Convolutional Networks (R-GCNs) to learn entity representations from knowledge graphs \cite{zhu2023dfmke} for item recommendations and utilize the Transformer framework to generate natural language responses. Based on the work of \citet{chen2019towards}, which mainly focuses on the entity-oriented knowledge graph, \citet{zhou2020improving} expand upon this by introducing an additional knowledge graph, i.e., the word-oriented knowledge graph -- ConceptNet \citep{speer2017conceptnet}. They propose a novel knowledge graph-based semantic fusion model for conversational recommender systems, harnessing the potential of both knowledge graphs. Subsequently, \citet{zhou2022c2} propose a contrastive learning approach\cite{won2023cross} from a coarse-to-fine perspective to enhance data semantic fusion. In order to mitigate redundant information in knowledge graphs, \citet{sarkar2020suggest} introduce different subgraphs to enhance recommendation performance for conversational recommender systems. To make use of multiple reasoning paths in knowledge graphs, \citet{CRWalker} and \citet{zhou2021crfr} propose conversational recommendation models to perform multi-hop reasoning in knowledge graphs to track shifts in users' interests. To alleviate the limitation of incomplete knowledge graphs, \citet{zhang2023variational} present a variational reasoning approach \citep{ren2022variational} with dynamic knowledge reasoning over incomplete knowledge graphs. Beyond knowledge graphs, pre-trained language models are also deployed to enhance conversational recommender systems \citep{wang2022towards, wang2022recindial, wang2023rethinking, xi2024memocrs, spurlock2024chatgpt, zhang2024improving, dao2024broadening, kemper2024retrieval, an2025beyond, wei2025mscrs}. \citet{wang2022recindial} finetune the large-scale pre-trained language models and propose a unified pre-trained language model-based framework to deal with the low-resource challenge in conversational recommender systems. Similarly, \citet{wang2022towards} apply a pre-trained language model and propose a conversational recommendation model on top of the prompt learning paradigm. \citet{dao2024broadening} introduce a demonstration-enhanced prompt learning method for conversational recommendation by retrieving and learning from demonstrations based on dialogue contexts. \citet{wei2025mscrs} propose a novel multi-modal semantic graph prompt learning framework for conversational recommender systems by considering collaborative information with multi-modal features. \citet{an2025beyond} propose a novel model of contextual disentanglement for conversational recommender systems by disentangling focus and background information from complex conversation contexts. 
Another promising direction in conversational recommendation involves the integration of side information \citep{lu2021revcore, yang2021improving}. For example, \citet{lu2021revcore} incorporate the item reviews to improve the recommendation performance in conversational recommender systems. \citet{li2022user} leverage multi-aspect user information to learn multi-aspect user preferences, while \citet{yang2021improving} introduce item meta information to improve item representations. 

Similar to the existing studies using side information, we also leverage the side information (e.g., item reviews and external knowledge graph) into our model. Different from the aforementioned research, we model multi-type data, including conversation history, knowledge graphs, and item reviews, through a unified mixture-of-experts framework to improve conversational recommender systems. 

\begin{figure}[t]
\centering
\includegraphics[width=\columnwidth]{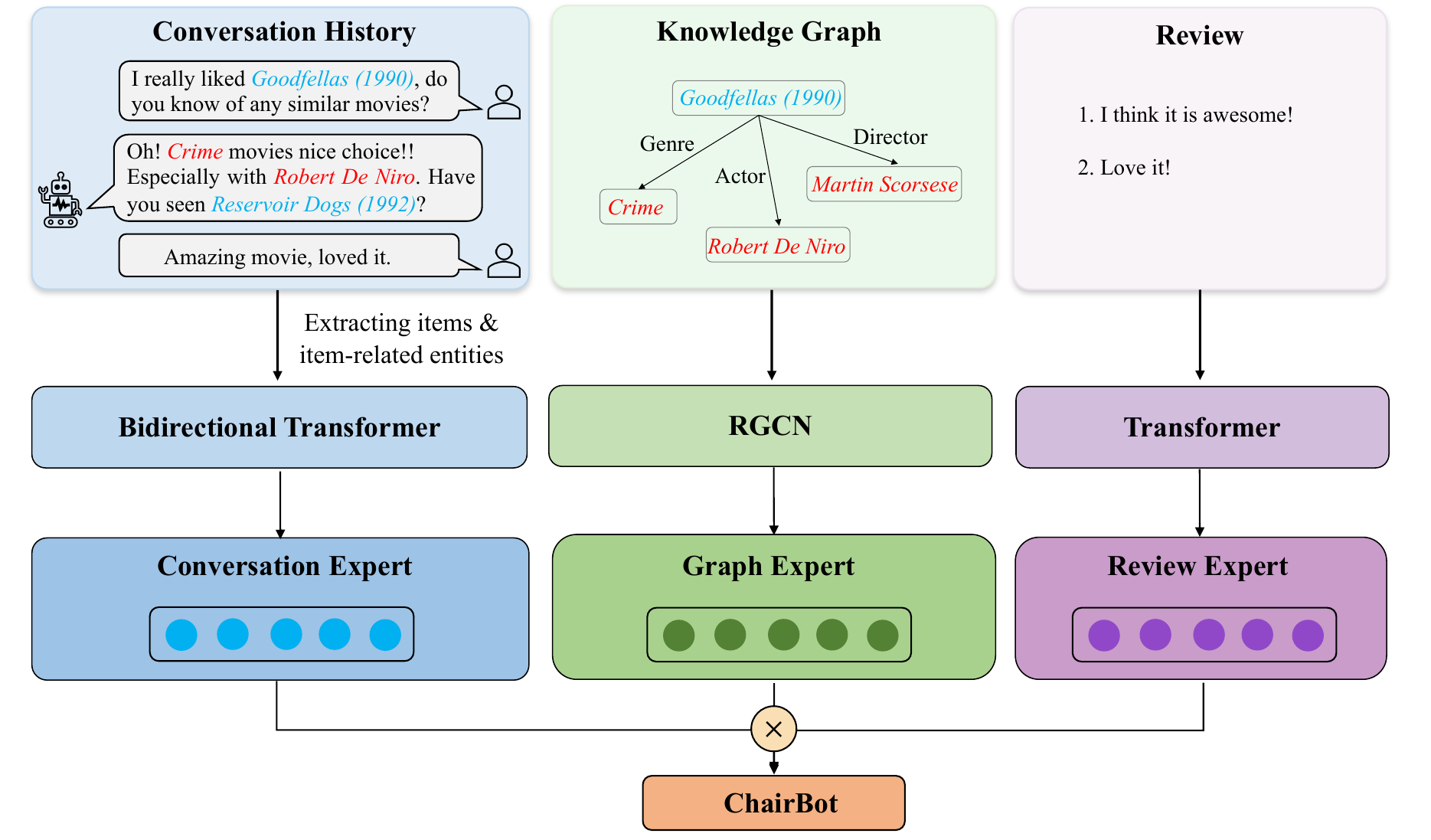}
\caption{The framework of our model in a movie recommendation scenario. Our model uses a ChairBot and multiple experts, including a conversation expert, a graph expert, and a review expert, to improve conversational recommender systems.}
\label{fig:method}
\end{figure}

\section{Approach}
In this section, we introduce the proposed MCCRS model. The proposed model contains a ChairBot and three experts, namely the conversation expert, the graph expert, and the review expert. Figure~\ref{fig:method} provides an overview of our model. In the following, we first describe the three experts in detail. Then, we describe our mixture-of-experts recommender to effectively fuse the three experts and our response generator benefiting from the recommender.

\subsection{Problem Formalization}
Formally, in conversational recommender systems, assume there is a user set $\mathcal{U}$ = $\{u_1, u_2, \dots, u_{|\mathcal{U}|}\}$, item set $\mathcal{I}$ = $\{i_1, i_2, \dots, i_{|\mathcal{I}|}\}$, and conversations. For a conversation, we have the contextual information, including conversation history consisting of a list of utterances, related reviews of item mentions in the conversation, and an external knowledge graph $\mathcal{G}$ (i.e., DBpedia). Each utterance is a sentence at $t$-th turn in the conversation, represented by a list of entity mentions (i.e., items or other item-related entities) in this paper. The entity set $\mathcal{E}$ = $\{e_1, e_2, \dots, e_{|\mathcal{E}|}\}$, consists of all the items and other item-related entities (i.e., $\mathcal{I} \subseteq \mathcal{E}$). Each review consists of a list of sentences $\mathcal{D}$ = $\{d_1, d_2, \dots, d_{|\mathcal{D}|}\}$. 
As the conversation goes on, at $t$-th turn, we aim to accurately recommend the mentioned item $i^*$ from the entire item set $\mathcal{I}$ based on the inferred user preference, along with a response, i.e., the next utterance comprises a sequence of words, to reply to the user. Table~\ref{table:Notations} summarizes the main notations used in the paper.

\begin{table}[t]
\caption{{Notations.}}
\label{table:Notations}
\begin{center}
\begin{tabular}{l|p{0.75\columnwidth}}
\toprule
Notation & Explanation\\
\hline
$u$, $\mathcal{U}$ & a user, and the user set\\
$i$, $\mathcal{I}$ & an item, and the item set\\
$e$, $\mathcal{E}$ & an entity, and the entity set\\
$r$, $\mathcal{R}$ & a relation, and the relation set\\
$d$, $\mathcal{D}$ & a review sentence, and the set of review sentences\\
$\mathbf{h}$, $\mathbf{H}$ & the elemental embedding and embedding matrix of the input sequences\\
$\mathbf{s}_k$, $\mathbf{p}_k$ & the  entity embedding and position embedding of $k$-th element in the input sequence\\
$\mathcal{G}$ & a knowledge graph \\
$\mathbf{D}_m$ & a representation matrix
for review sentences\\
$\mathbf{v}_R$ & a review embedding \\
$C$, $G$, $R$ & the conversation expert, graph expert, and review expert\\
$\mathbf{n}_{e}^{(l)}$  & the representation of node (entity) $e$ at the $l$-th layer of R-GCN\\
$\lambda$  & the importance score of experts\\
\bottomrule
\end{tabular}
\end{center}
\end{table}

\subsection{Conversation Expert}
The conversation expert is made up of a sequential transformer \cite{zou2022improving-sigir}. It extracts both the mentioned items and item-related entities in conversations to form a user sequence. For instance, in Figure~\ref{fig:method}, a sequence of [Goodfellas (1990), Crime, Robert De Niro, Reservoir Dogs (1992)] is extracted for the conversation history. 
Based on this user sequence, the conversation expert applies a Cloze task~\citep{devlin2018bert, taylor1953cloze} to randomly mask a portion of the sequences and puts them into a transformer to predict the masked items. The transformer is uniquely designed, with a matrix of hidden size $\times$ vocabulary size in place of the usual positional encoding. The presentation of $k$-th element of the input sequence is the concentration of positional embedding $\mathbf{s}_k$ and sequence embedding $\mathbf{p}_k$.
\begin{equation}
\mathbf{h}_k=\mathbf{s}_k+\mathbf{p}_k.
\end{equation}

All elemental embeddings of $\mathbf{h}_k$ from the previous steps form a matrix $\mathbf{H}$, which will then be passed through a multi-attention head. Suppose we are at the $n$-th transformer layer, we will update $\mathbf{H}$ in the following manner:
\begin{equation}
\mathbf{H}^{n+1}=\text{MultiHead}(\text{PFFN}(\mathbf{H}^{n})),
\label{equ:1}
\end{equation}
where MultiHead is a multi-head self-attention sub-layer, and PFFN is a Position-wise Feed-Forward Network constructed by the Feed-Forward Network (FFN) with GELU activation \citep{hendrycks2016bridging}.

\jie{Afterward, we would combine their calculation with a residual link. The reason for the residual connection is that the network is more difficult to train as it becomes deeper, and thus employing such a technique, along with layer normalization and dropout, assists with the training.}
\begin{equation}
\mathbf{H}^{n}=\text{LayerNorm}(\mathbf{H}^{n} + \text{Dropout}(\text{sublayer}(\mathbf{H}^{n})),
\end{equation}
\jie{where the sublayer is either PFFN or MultiHead in Equation \ref{equ:1}.}

Assuming we are predicting $k$-th movie, we utilize the final embedding at N-layer of Transformer $\mathbf{h}^{N}_k$ to generate our prediction. 
More specifically, we utilize a softmax function to output the item distribution:
\begin{equation}
P_C(i)=\text{Softmax}(\mathbf{W} \times \mathbf{h}^{N}+\mathbf{b}),
\end{equation}
where $\mathbf{W}$ is a learnable transformation matrix, and $\mathbf{b}$ denotes the trainable bias matrix. 
On top of the produced probability of items, we use a cross-entropy loss to optimize the parameters of the conversation expert. 
\begin{equation}
\label{equ:loss}
L_{rec}= - \sum_{j=1} \sum_{i \in \mathcal{I} } y_{ij} \cdot \log P(i),
\end{equation}
\jie{where $j$ is the conversation index, and $y_{ij}$ indicates the ground-truth label of items. After the training, we collect both the predicted item probability $P_C(i)$ and hidden embedding $\mathbf{h}^{N}$ for the ChairBot.}

\begin{figure}[t]
\centering
\includegraphics[width=0.6\columnwidth]{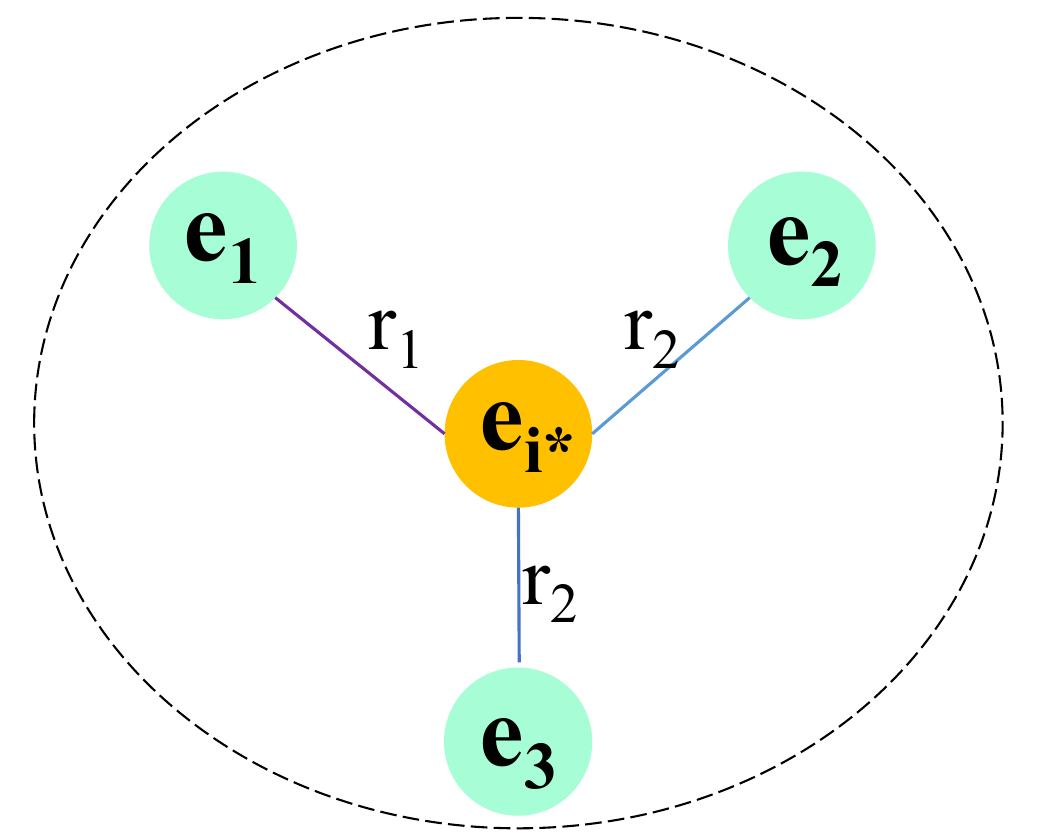}
\caption{Illustration of an example graph.}
\label{Graph}
\end{figure}
\subsection{Graph Expert}

Given that it is difficult to comprehensively understand user preferences based solely on conversational context, the inclusion of external knowledge is necessary to encode user preferences. For dialogs in conversational recommender systems, item mentions and item-related entities can be extracted to construct external knowledge graphs. Inspired by the previous studies \citep{chen2019towards, zhou2020improving}, we introduce a
knowledge graph sourced from DBpedia \citep{lehmann2015dbpedia} to encode structural and relational information in the knowledge graph. Specifically, we perform entity linking \citep{ferragina2010tagme} to map the item mentions and item-related entities in the dataset to DBpedia. With the help of the external knowledge graph, it enables us to model user preferences more accurately. 

A knowledge graph $\mathcal{G}$ (i.e., DBpedia) comprises an entity set $\mathcal{E}$ and a relation set $\mathcal{R}$. The knowledge graph $\mathcal{G}$ stores semantic facts as triples $<e_1, r, e_2>$, where $e_1, e_2 \in \mathcal{E}$ represents items or item-related entities and $r \in \mathcal{R}$ represents the relation between $e_1$ and $e_2$. 
For example, as shown in Figure~\ref{Graph}, for a target movie $e_{i^*}$, it might be associated with entity $e_1$ with one type of relationship, while $e_2$ and $e_3$ are of another, such as $e_1$ being the genre while $e_2$ and $e_3$ are the actors in the movies. 

\jie{In this paper, we employ R-GCN \citep{schlichtkrull2018modeling} to encode entity representations in the knowledge graph $\mathcal{G}$. Specifically, we pre-train the representation $e \in \mathcal{E}$ by using R-GCN to initialize the offline embeddings of items or other item-related entities. Formally, in R-GCN, the node embedding at $(l + 1)$-th layer is calculated as:}
\begin{equation}
\label{equ:rgcn}
\mathbf{n}_{e}^{(l+1)}=\sigma(\sum_{r \in \mathcal{R}} \sum_{e^{\prime} \in \mathcal{E}_{e}^{r}} \frac{1}{Z_{e, r}} \mathbf{W}_{r}^{(l)} \mathbf{n}_{e^{\prime}}^{(l)}+\mathbf{W}^{(l)} \mathbf{n}_{e}^{(l)}),
\end{equation}
where $\mathbf{n}_{e}^{(l)}$ denotes the representation of node (i.e., entity) $e$ at the $l$-th layer, and $\mathcal{E}_{e}^{r}$ is the set of neighboring nodes for $e$ under the relation $r$. $\mathbf{W}^{(l)}$ represents a learnable transformation matrix for transforming the representations of nodes at the $l$-th layer $\mathbf{n}_{e}^{(l)}$, while $\mathbf{W}_{r}^{(l)}$ is another relation-specific learnable matrix for transforming the embedding of neighboring nodes under the relation $r$. $\sigma$ is the ReLU activation function and $Z_{e, r}$ denotes a normalization factor.

After aggregating the structural and relational information of the knowledge graph, we obtain all the node representations (i.e., representations of items and other item-related entities) on the top R-GCN layer. As historically interacted entities are critical for modeling user preference, we aggregate the overall representation for each conversation. Specifically, we obtain the entity mentions for each conversation, and then apply a self-attention mechanism, that can automatically consider the levels of entity importance, to obtain the graph-based user representation $\mathbf{n_{e_u}}$. We compute the item probability to rank all the items:
\begin{equation}
P_G(i)=\text{Softmax}({\mathbf{n}^\intercal_{e_u}} \mathbf{n}_{i}),
\label{eq:softmaxkg}
\end{equation}
\jie{where $\mathbf{n}_{i}$ represents the learned embedding for item $i$. Again, a cross-entropy loss is used for the graph expert similar to the conversation expert (i.e., Equation~\ref{equ:loss}). Finally, the graph expert produces the item probability $P_G(i)$ and node embeddings for the ChairBot.}

\subsection{Review Expert}
Besides the conversation history and knowledge graph, the review data related to items is able to improve the performance of conversational recommender systems. The review data is a set of sentences about items written by online users. To encode review text, we apply a two-step approach. First, we encode each sentence by employing the standard Transformer model. 
Second, we utilize a sentence-level self-attention layer to produce the overall review-based representation. For item $i$, assuming it has $m$ sentences, we first obtain $\mathbf{D^m}$, which is a representation matrix for sentences where each column is a sentence representation~\cite{schlichtkrull2018modeling}. Then the overall review-based representation is:
\begin{equation}
\mathbf{v}_R=\text{SelfAttention}(\text{Transformer}(\mathbf{D^m})).
\end{equation}

After the above encoding, we pass $\mathbf{v}_R$ through another self-attention layer to produce the final review embeddings.\footnote{We notice that the review data is missing for a few movie entities. In this case, we take the mean of the review representation matrix as the representation of the missing values. One can also use other techniques to deal with the missing reviews.} We generate the item probability, by using the final review embeddings to replace the node embeddings in Equation~\ref{eq:softmaxkg}: 
\begin{equation}
P_R(i)=\text{Softmax}({\mathbf{v}^\intercal_{R_u}} \mathbf{v}_{R_i}).
\label{eq:softmaxreview}
\end{equation}

Also, a cross-entropy loss is used for optimizing the review expert similar to the conversation expert (i.e., Equation~\ref{equ:loss}). \jie{Again, the item probability $P_R(i)$ and review embeddings are passed to the ChairBot. }

\subsection{Mixture-of-Experts Recommender}
After the above three experts, which are specialized in three particular types of data, we then fuse them by introducing a ChairBot. The ChairBot coordinates the three experts and integrates their results to generate more accurate and relevant recommendations. Inspired by LSTM-like structure \cite{moe}, we first collect the hidden embeddings from all experts $\mathbf{h}_i^C, \mathbf{h}_i^G, \mathbf{h}_i^R$, and the predictions from all experts $\mathbf{p}_i^C, \mathbf{p}_i^G, \mathbf{p}_i^R$ for the item $i$, and concatenate them. The $C$, $G$, and $R$ denote conversation expert, graph expert, and review expert respectively.
\begin{equation}
\begin{aligned}
\mathbf{h}_C &=\mathbf{h}_i^C \oplus \mathbf{p}_i^C,\\
\mathbf{h}_G &=\mathbf{h}_i^G \oplus \mathbf{p}_i^G,\\
\mathbf{h}_R &=\mathbf{h}_i^R \oplus \mathbf{p}_i^R.\\
\end{aligned}
\end{equation}

Then, we generate the normalized importance score:
\begin{equation}
\begin{aligned}
\beta_b=\textit{MLP}(\mathbf{h}_b),\\
\lambda_b=\frac{\beta_b}{\beta_C+\beta_G+\beta_R},
\end{aligned}
\end{equation}
where ${b ={C,G,R}}$, and $\textit{MLP}$ is a linear layer.

Finally, we compute the recommendation probability for an item:
\begin{equation}
P_{rec}(i)= \lambda_C * P_C(i) + \lambda_G * P_G(i) +\lambda_R * P_R(i),
\end{equation}
\jie{where $P_C(i)$, $P_G(i)$, and $P_R(i)$ are item probabilities generated by the conversation expert, graph expert, and review expert, respectively.} To fine-tune the entire item recommender, we also apply a cross-entropy loss in the Mixture-of-Experts recommender module (i.e., Equation~\ref{equ:loss}) to fine-tune the representations and improve the performance.\footnote{One can also use the joint loss for all experts and ChairBot to optimize the recommender. We decided to train all experts separately because we observed better performance in the experiments when training them separately.}

\subsection{Response Generator}
The response generator can be improved by the recommender system as it provides recommendation-aware vocabulary bias. As such, the recommender system helps generate more consistent and diverse responses. Following \citet{zhou2020improving}, we integrate multiple cross-attention layers within a standard Transformer decoder architecture to merge the pre-trained representations effectively.  
We modify the decoder by fusing the entity representations from our conversation expert, graph expert, and review expert:
\begin{equation}
\begin{aligned}
\mathbf{A}^l_0 &=\text{MultiHead}[\mathbf{B}^{(l-1)},\mathbf{B}^{(l-1)},\mathbf{B}^{(l-1)}],\\
\mathbf{A}^l_1 &=\text{MultiHead}[\mathbf{A}^l_0,\mathbf{F}_C,\mathbf{F}_C],\\
\mathbf{A}^l_2 &=\text{MultiHead}[\mathbf{A}^l_1,\mathbf{F}_G,\mathbf{F}_G],\\
\mathbf{A}^l_3 &=\text{MultiHead}[\mathbf{A}^l_2,\mathbf{F}_R,\mathbf{F}_R],\\
\mathbf{A}^l_4 &=\text{MultiHead}[\mathbf{A}^l_3,\mathbf{X},\mathbf{X}],\\
\mathbf{B}^{l} &=\text{FFN}(\mathbf{A}^l_4),
\end{aligned}
\end{equation}
where MultiHead[·, ·, ·] represents the multi-head attention function. FFN(·) denotes a feed-forward network. $\mathbf{X}$ is the representation matrix of dialogue history obtained from a standard Transformer encoder. $\mathbf{B}^{l}$ is the representation matrix at the $l$-th layer from the decoder. $\mathbf{A}^l_0$, $\mathbf{A}^l_1$, $\mathbf{A}^l_2$, $\mathbf{A}^l_3$, and $\mathbf{A}^l_4$ are embeddings after self-attention, cross-attention with $\mathbf{F}_C$ from the conversation expert, cross-attention with $\mathbf{F}_G$ from the graph expert, cross-attention with $\mathbf{F}_R$ from the review expert, and cross-attention with the encoder output $\mathbf{X}$, respectively.

\section{Experiments}
In our experiments, we want to address the following research questions: 

\RQ{1}{How does our proposed model perform in comparison to prior baselines?}
\RQ{2}{What is the impact of different components of the model?}
\RQ{3}{How do various parameters of our proposed model influence its effectiveness?}

\subsection{Experimental Setup}

\begin{table}[t]
\caption{The dataset statistics in our experiments.}
\label{table:data}
\centering
\small
\begin{tabular}{lcccc}
\toprule
\textbf{Dataset} & \textbf{\# Conversations} & \textbf{\# Utterances} & \textbf{\# Users} & \textbf{\# Items}\\
\midrule
INSPIRED & 1,001 & 35,811 & 1,482 & 1,783\\
ReDial & 10,006 & 182,150 & 956 & 51,699\\
\bottomrule
\end{tabular}
\end{table}
\subsubsection{Dataset}
We use the two typical conversational recommendation datasets, ReDial \citep{li2018towards} and INSPIRED \citep{hayati2020inspired}, as done in \citet{wang2022towards}. 
The dataset statistics are presented in Table \ref{table:data}. ReDial \citep{li2018towards} is an English dataset for conversational recommendation, including a collection of annotated dialogs where a recommender offers movie suggestions for the seeker. This dataset contains 10,006 conversations, 182,150 utterances, 956 users, and 51,699 movies. INSPIRED is another dataset of conversational movie recommendations, though it is smaller in scale compared to ReDial. Both datasets are commonly used to evaluate conversational recommender systems \citep{wang2022towards}. 
The datasets are split into training, validation, and test sets by 8:1:1 ratio by prior studies \cite{li2018towards, wang2022towards}. For review data that is not contained in the above datasets, we retrieve reviews for movies from IMDb~\footnote{https://www.dbpedia.org/} similar to \citet{zhou2022c2}. 

\subsubsection{Evaluation Metrics}
For evaluating recommendations in conversational recommender systems, we utilize Recall@k (where k = 1, 10, and 50) as our metrics, following previous work\citep{chen2019towards, zhou2020improving, zhou2022c2}. 
In each conversation, we begin evaluation from the first sentence of the system’s responses, i.e., we treat each item or utterance from the recommender as ground truth and assess them sequentially throughout the conversation, consistent with prior work~\citep{chen2019towards, zhou2020improving}. When testing, we use all items as candidates to rank for recommendations.

For evaluating the conversation quality, we use both automated and human evaluation methods. For automated evaluation, we apply Distinct-n (n = 2, 3, 4) to measure sentence-level diversity, as done in previous studies \citep{ren2022variational, zhou2022c2}. 
For human evaluation, three annotators are invited to review the generated responses of all baselines and our model on two dimensions: fluency and informativeness. Fluency assesses the natural flow of the responses, while informativeness measures the relevance and interest of the information provided. The annotators rate the responses on a scale from 0 to 2, and the average score from all three annotators is calculated to report the results.

\subsubsection{Implementation Details}
The model is trained via the Adam optimizer \citep{kingma2014adam} with a batch size of 256. The learning rate is 1e-4. We use 50 as the maximum length of sequence K. To ensure training stability, a gradient global norm clipping of 5 and an L2 regularization of 0.01 are adopted. 
For the Transformer, we set the number of Transformer layers N as 2 and the head number as 2. 
The number of R-GCN layers and the normalization factor $Z_{e, r}$ of R-GCN are set to 1. 
The parameters of mask proportion and hidden dimensionality are discussed in Section \ref{subsec:para}. 

\begin{table}[t]
\centering
\caption{Recommendation results on ReDial. `*' indicates significant improvements over the best baseline (Fisher random test, $p$-value $<0.05$). The best performances are highlighted in bold, while the second-best are underlined (Unless otherwise reported, we use `*' to indicate significant improvements, bold values to indicate the best performances, and underlined values to indicate the second-best performances throughout the paper). Our proposed MCCRS method significantly outperforms SOTA baselines on the recommendation task on ReDial.}
\label{tab:baseline}
\begin{tabular}{lccc}
  \toprule
  \textbf{Model} & \textbf{Recall@1} & \textbf{Recall@10} & \textbf{Recall@50} \\ 
  \midrule
    Popularity & 0.012 & 0.061 & 0.179\\
    TextCNN & 0.013 & 0.068 & 0.191\\
    BERT & 0.014 & 0.117 & 0.191\\
    ReDial & 0.023 & 0.129 &0.287\\
    KBRD & 0.031 & 0.150 & 0.336\\
    KGSF & 0.039 & 0.183 & 0.378\\
    RevCore & 0.046 & 0.220 & 0.396\\
    VRICR & \underline{0.054} & \underline{0.244} & 0.406\\
    $\text{C}^2$-CRS & 0.053 & 0.233 & \underline{0.407}\\ 

  \midrule
  MCCRS & \textbf{0.057*} &	\textbf{0.250*} &	\textbf{0.473*}\\ 
  \bottomrule
\end{tabular}
\end{table}
\subsubsection{Baselines}
In this work, we compare our approach with four typical baselines, along with several representative baselines commonly used on ReDial and INSPIRED.
\begin{itemize}
    \item \textbf{Popularity} is a typical approach ranking the items based on historical recommendation frequency. 
    \item \textbf{TextCNN} \citep{kim2014convolutional} is a typical model that is based on CNN representations.
    \item \textbf{BERT} \citep{devlin2018bert} is a typical method, which learns from the sentences to generate recommendations.
    \item \textbf{Transformer} \citep{vaswani2017attention} is a typical encoder-decoder model based on the Transformer architecture, widely used for generating responses.
    \item \textbf{ReDial} \citep{li2018towards} serves as a benchmark model for ReDial by employing an autoencoder-based recommender system. 
    \item \textbf{KBRD} \citep{chen2019towards} leverages the DBpedia knowledge graph and incorporates a knowledge-enhanced recommender to improve conversational recommendation. 
    \item \textbf{KGSF} \citep{zhou2020improving} incorporating word-oriented and entity-oriented knowledge graphs for the conversational recommendation. 
    \item \textbf{RevCore}~\citep{lu2021revcore} is a review-augmented conversational recommender by leveraging reviews to enrich item information. 
    \item \textbf{VRICR}~\citep{zhang2023variational} is one of the SOTA conversational recommendation methods, which is based on a variational reasoning method to complete the missing information in incomplete knowledge graphs.
    \item \textbf{$\text{C}^2$-CRS}~\citep{zhou2022c2} is one of the SOTA conversational recommendation methods that utilizes contrastive learning and semantic fusion of multi-type information including context sentences, reviews, and knowledge graphs. 
\end{itemize}
Among these baselines, Popularity, TextCNN, BERT, and Transformer are classical methods, while ReDial, KBRD, KGSF, RevCore, $\text{C}^2$-CRS, and VRICR 
are conversational recommendation methods. We do not compare with conversational recommendation methods which use large pre-trained language models due to different task settings. For parameters used in baselines, we utilize the optimal values as reported in the respective papers. 

\subsection{Overall Performance (RQ1)}

\begin{table}[tb]
\centering
\caption{Recommendation results on INSPIRED. 
Our proposed MCCRS method significantly outperforms SOTA baselines on the recommendation task on INSPIRED. }
\label{tab:baseline2}
\begin{tabular}{lccc} 
  \toprule
  \textbf{Model} & \textbf{Recall@1} & \textbf{Recall@10} & \textbf{Recall@50} \\ 
  \midrule
    Popularity & 0.032 & 0.155 & 0.323\\
    TextCNN & 0.025 & 0.119 & 0.245\\
    BERT & 0.044 & 0.179 & 0.328\\
    ReDial & 0.031 & 0.117 & 0.285\\
    KBRD & 0.058 & 0.146 & 0.207\\
    KGSF & 0.058 & 0.165 & 0.256\\
    RevCore & 0.068& 0.198& 0.379\\
    VRICR & 0.043 & 0.141 & 0.336\\
    $\text{C}^2$-CRS & \underline{0.090} & \underline{0.242} & \underline{0.399}\\  
  \midrule
  MCCRS & \textbf{0.104*} & \textbf{0.275*} &	\textbf{0.497*}\\
  \bottomrule
\end{tabular}
\end{table}

\subsubsection{Performance on Recommendation Task} 
In this section, we explore the effectiveness of our proposed method in comparison to SOTA baselines. We evaluate our recommendation performance against baselines on ReDial and INSPIRED, as shown in Table~\ref{tab:baseline} and Table~\ref{tab:baseline2}, respectively. 
For the evaluation metrics, the average performance across the maximum number of conversational turns is reported.

For the recommendation performances on ReDial in Table~\ref{tab:baseline}, we observe that our proposed MCCRS significantly outperforms all the baselines on all metrics on the ReDial dataset. MCCRS outperforms the best classical recommendation baseline BERT by 307\%, 114\%, and 148\% in terms of Recall@1, Recall@10, and Recall@50, respectively. Compared with the SOTA conversational recommendation baselines $\text{C}^2$-CRS and VRICR, MCCRS outperforms $\text{C}^2$-CRS by 7.5\%, 7.3\%, and 16.2\% while outperforms VRICR by 5.6\%, 2.5\%, and 16.5\%, 
in terms of Recall@1, Recall@10, and Recall@50, respectively. 
Similarly, observe from Table~\ref{tab:baseline2} for the recommendation performances on INSPIRED, our proposed model, MCCRS, still significantly outperforms all the baselines on all metrics. MCCRS achieves a significant improvement of 136.4\% on Recall@1, 53.6\% on Recall@10, and 51.5\% on Recall@50 over the best classical baseline BERT. Compared with the SOTA conversational recommendation baseline $\text{C}^2$-CRS, MCCRS achieves a significant improvement of 15.6\% on Recall@1, 13.6\% on Recall@10, and 24.6\% on Recall@50. The above observations indicate the effectiveness of our MCCRS model by incorporating multiple experts. MCCRS utilizes multi-type external information, including conversational history, knowledge graph, and reviews, which is helpful in understanding the conversation context. The effective Mixture-of-Experts framework to fuse the multi-type external information enhances the data representations and improves the recommendation performance. For conversational baselines, we find that 
KBRD, and KGSF perform better than ReDial, due to their use of external knowledge graphs, which aid in interpreting user intentions more effectively. RevCore achieves better recommendation performance than KBRD and KGSF. This might be because RevCore incorporates additional review information on items. $\text{C}^2$-CRS outperforms RevCore, as it performs semantic fusing techniques with more external information.

\begin{table}[t]
\centering
\caption{Response generation results on ReDial (automatic evaluation). 
Our proposed MCCRS method achieves better performance of response generation than baselines on the ReDial dataset. }
\label{tab:Autoconv}
\begin{tabular}{lccc} 
  \toprule
  \textbf{Model} & \textbf{Distinct-2} & \textbf{Distinct-3} & \textbf{Distinct-4} \\ 
  \midrule
  Transformer & 0.148 &  0.151 &  0.137 \\
  ReDial & 0.225 & 0.236 & 0.228 \\
  KBRD &  0.263 &  0.368 &  0.423 \\
  KGSF & 0.330 &  0.417 &  0.521 \\
  RevCore & 0.424 & 0.558 &  0.612\\
   VRICR & 0.382 & 0.453 & 0.496\\
  $\text{C}^2$-CRS &  \underline{0.631} & \underline{0.932} & \underline{0.909}  \\
  \midrule
  MCCRS & \textbf{0.680*} & \textbf{0.976*} & \textbf{0.981*}\\ 
  \bottomrule
\end{tabular}
\end{table}

\begin{table}[t]
\centering
\caption{Response generation results on INSPIRED (automatic evaluation). 
Our proposed MCCRS method achieves better performance of response generation than baselines on the INSPIRED dataset. }
\label{tab:Autoconv2}
\begin{tabular}{lccc} 
  \toprule
  \textbf{Model} & \textbf{Distinct-2} & \textbf{Distinct-3} & \textbf{Distinct-4} \\ 
  \midrule
  Transformer &  1.020 & 2.248 & 3.582 \\
  ReDial &  1.347 &  1.521 & 3.445 \\
  KBRD & 1.369 & 2.259 &  3.592 \\
  KGSF & 1.608 & 2.719 & 4.929 \\
  RevCore & 2.419 & 3.820 & 4.648\\
  VRICR & 1.937 & 3.248 & 4.965\\
  $\text{C}^2$-CRS & \underline{2.456} & \underline{4.432} & \underline{5.092} \\
  \midrule
  MCCRS & \textbf{2.584*} & \textbf{4.579*} & \textbf{5.251*}\\  
  \bottomrule
\end{tabular}
\end{table}

\subsubsection{Performance on Conversation Task} 
In this subsection, we verify the performance of our proposed model on the conversation task. Specifically, we report the evaluation metrics for both automatic and human evaluation.

\paragraph{Automatic Evaluation}
Table~\ref{tab:Autoconv} and Table~\ref{tab:Autoconv2} show the performance comparison of automatic evaluation on response generation, on ReDial and INSPIRED, respectively. Among baselines, we observe that ReDial, KBRD, and KGSF achieve better performance than Transformer on both ReDial and INSPIRED, as they apply pre-training or enhance word probability by leveraging items and item-related entities from external knowledge graphs. $\text{C}^2$-CRS performs the best among baselines. One possible reason is that $\text{C}^2$-CRS applies a semantic fusion approach and an instance weighting mechanism to improve the diversity of responses. 

Compared with these baselines, we see that our MCCRS model outperforms all baselines in terms of automatic metrics (i.e., Distinct-2, Distinct-3, and  Distinct-4) on both ReDial and INSPIRED datasets. This might be because MCCRS effectively leverages multi-type information to understand the context and against noisy information. The multi-type context-aware framework enhances the diversity of generated responses.

\begin{table}[tb]
\centering
\caption{Response generation results on ReDial (human evaluation). 
Our proposed MCCRS method generates more fluent and informative responses than baselines on the conversation task. }
\label{tab:humanconv}
\begin{tabular}{lcc} 
  \toprule
  \textbf{Model} & \textbf{Fluency} & \textbf{Informativeness} \\ 
  \midrule
  Transformer & 0.82  & 0.91 \\
  ReDial & 1.25 & 1.09 \\
  KBRD  & 1.31 & 1.22 \\
  KGSF & 1.53 & 1.32 \\
  RevCore & 1.55 & 1.38 \\
  VRICR  &1.52 & 1.34 \\
  $\text{C}^2$-CRS  & \underline{1.58} & \underline{1.51}\\
  \midrule
   MCCRS & \textbf{1.66*} & \textbf{1.59*}\\
  \bottomrule
\end{tabular}

\end{table}
\begin{table}[tb]
\centering
\caption{Results of ablation study for recommendation. Removing any of the three experts lowers the recommendation performance.}
\label{tab:Ablation}
\begin{tabular}{lccc} 
  \toprule
  \textbf{Model} & \textbf{Recall@1} & \textbf{Recall@10} & \textbf{Recall@50} \\ 
  \midrule
  Conversation Only& 0.022 & 0.113 & 0.207\\
  Graph Only& 0.051 & 0.213 & 0.373\\
  Review Only	& 0.021 & 0.093& 0.369\\
  w/o conversation& 0.053 & 0.216 & 0.471\\
  w/o graph& 0.025 & 0.121 & 0.382\\
  w/o review	& 0.052 & 0.228 & 0.472\\ 
  \midrule
  MCCRS & \textbf{0.057*} &	\textbf{0.250*} &	\textbf{0.473*}\\  
  \bottomrule
\end{tabular}
\end{table}
\paragraph{Human Evaluation}
Table~\ref{tab:humanconv} shows the performance comparison of human evaluation on response generation. From the results, we find that ReDial outperforms the typical Transformer by incorporating a pre-trained RNN encoder. $\text{C}^2$-CRS performs the best among baselines on the reported human evaluation metrics. \jie{This might be because $\text{C}^2$-CRS facilitates semantic fusion of data by using contrastive learning, which improves the responses.} 

Compared with these baselines, we observe that our MCCRS method performs better than all baselines in terms of the reported human evaluation metrics, i.e., fluency and informativeness. This demonstrates that our MCCRS model can generate more informative words or entities while maintaining the fluency of the generated responses. This might be because our MCCRS model incorporates multi-type external information to generate fluent responses. The high accuracy of the recommender module leads to generating high-quality items, which facilitates the informativeness of responses. 

\subsection{Ablation Study (RQ2)}
We explore the contributions of different components within our model in this section. We perform an ablation study on the ReDial dataset, by considering six variants of the model, including (1) ``w/o conversation'' removes the conversation expert from the framework; (2) ``w/o graph'' removes the graph expert;  (3) ``w/o review'' removes the review expert;  (4) ``conversation only'' only involves the single conversation expert; (5) ``graph only'' only involves the single graph expert; (6) ``review only'' only involves the single review expert. 
Observe from Table~\ref{tab:Ablation}, among the three experts, ``graph only'' achieves the highest performance than the other two single experts. \jie{Also, we find that removing any of the experts, either the conversation expert, graph expert, or review expert, results in lower performance. This indicates that all three experts contributed to the final performance.} After removing the graph expert, the performance drops more than the other two experts. This indicates the graph expert contributes the most among the three experts, highlighting the importance of the graph expert. This might be because the graph expert benefits from the structure of the knowledge graph. Moreover, although we observe that ``review only'' and ``conversation only'' achieve relatively low performance, removing the conversation expert or review expert results in a decrease in performance. This suggests that adding conversations or reviews will marginally improve the recommendation accuracy. The reviews, although did not play a major role, do help with the recommendation task even if reviews are missing for some items.

\subsection{Parameter Sensitivity (RQ3)}
\label{subsec:para}
\jie{In this section, we explore how our proposed model is affected by its main parameters, including the mask proportion and hidden dimensionality, on the ReDial dataset.}

\subsubsection{Mask Probability}
When modeling the conversation, we apply a Cloze task to randomly mask a portion of the sequences. We examine the impact of mask probability \citep{zou2022improving-sigir} on model performance. Observe from Table \ref{tab:MaskProb}, performance initially improves as the mask proportion increases, but then declines at higher mask proportions. We conclude that mask proportion indeed affects performance. The optimal mask proportion is 0.4. 

\subsubsection{Hidden Dimensionality}
Table \ref{tab:dimension} illustrates the effect of hidden dimensionality (embedding size) on model performance. Overall, the recommendation performance generally declines as dimensionality increases from 32 to 256. This is probably due to overfitting. The proposed model achieves its best performance with a hidden dimension of 32.

\begin{table}[tb]
\centering
\caption{Effect of Mask Probability.}
\label{tab:MaskProb}
\begin{tabular}{cccc} 
  \toprule
  \textbf{Mask Probability} & \textbf{Recall@1} & \textbf{Recall@10} & \textbf{Recall@50}  \\ 
  \midrule
  0.2 &  0.044 & 0.211 & 0.419 \\
  0.4	& 0.057	& 0.250	& 0.473\\
  0.6	& 0.041 & 0.209 & 0.413 \\
  0.8 & 0.047 & 0.220 & 0.433 \\
  \bottomrule
\end{tabular}
\end{table}

\begin{table}[tb]
\centering
\caption{Effect of hidden dimensionality.}
\label{tab:dimension}
\begin{tabular}{cccc} 
  \toprule
  \textbf{Hidden Dimensionality} & \textbf{Recall@1} & \textbf{Recall@10} & \textbf{Recall@50}  \\ 
  \midrule 
  32 & 0.057 & 0.250	& 0.473\\
  64 & 0.043&0.215&0.379\\
  128	& 0.046 & 0.220	& 0.444 \\
  256 & 0.038 & 0.174 & 0.335 \\
  \bottomrule
\end{tabular}
\end{table}
\section{Conclusion and Future Work}
In this paper, we proposed the MCCRS model for conversational recommender systems. The MCCRS model deploys a mixture-of-experts structure to leverage multi-type data, including the structured knowledge graph, unstructured conversation history, and unstructured item reviews. It consists of three experts, with each expert specialized in a particular type of contextual data. Multiple experts are then coordinated by a ChairBot by considering the importance of each expert to generate recommendations. The embeddings from the three experts are introduced to the response generator as cross-attention embeddings to generate responses. The experimental results on two widely used datasets demonstrate that our MCCRS model outperforms the SOTA conversational recommendation baselines and is highly effective. 

One limitation of this work is that we use entities from the DBpedia knowledge graph following previous work of conversational recommender system~\citep{chen2019towards,sarkar2020suggest,zhou2020improving,CRWalker}, the extracted entities may not be 100\% accurate. Therefore, it is valuable to explore the conversational recommender system by integrating and modeling the uncertainty and noise in conversation contexts. \jie{ Also, we do not incorporate sentiment analysis of items, entities, or reviews in the multi-type contextual data. In future work, we plan to explore conversational recommender systems by integrating and modeling the sentiment information (e.g., positive or negative) from the conversation and reviews in future work.} 

Also, for each conversation, we utilize all the data from multi-type resources. Although the experimental results validate its effectiveness, it is worth detecting the consistency of different types of data sources in conversational recommender systems, enabling the conversational recommendation model to predict the necessity for integrating each type of contextual data and model the inconsistency of different types of contextual data. Furthermore, we model the conversation by utilizing the full sequence. However, the inherent complexity of conversations, exemplified by scenarios such as multi-topic discussions where users seamlessly transition between different subjects, poses a challenge as these conversations may lack strict sequential dependencies within the full sequence. A potential research direction is to partition the entire conversation sequence into discrete subsequences (e.g., detecting topic threads in multi-topic conversations \cite{zhou2020towards} and modeling each topic as a subsequence) so that more strict sequential dependencies remain in subsequences. One can also model the noise in the sequence to relax the strict sequential dependencies.

\jie{Last, while MCCRS achieves strong performance in benchmark datasets and effectively fuses multi-type contextual information, its mixture-of-experts structure may lead to increased computational complexity, which may pose challenges in real-world deployment scenarios when involving large-scale or real-time recommendation tasks. In future work, we plan to explore potential techniques, such as knowledge distillation for model compression \citep{11072408} and expert pruning, for improving model efficiency and scalability. Also, for the mixture-of-experts structure, we combine the results of multiple experts with a followed ChairBot. One can also use other combination strategies of multiple experts, e.g., attention-based gating mechanisms, or a hierarchical structure that sequentially refines expert outputs. }

\section{Acknowledgement}
We would like to thank Qingtian Cao for her contribution to the early exploration and discussion of this work. This research was supported by the National Natural Science Foundation of China (62402093) and, the Sichuan Science and Technology Program (2025ZNSFSC0479). This work was also supported in part by the National Natural Science Foundation of China under grants U20B2063 and 62220106008, and the Sichuan Science and Technology Program under Grant 2024NSFTD0034.

\bibliographystyle{elsarticle-num-names} 
\bibliography{cas-refs}

\end{document}